# When the Algorithm Becomes the Brand Crisis:

## A Sociotechnical Theory of Distributed Responsibility and Accountable Transparency

Mohammad Saleh Torkestani
Senior Lecturer in Marketing
University of Exeter, UK
Email: m.torkestani@exeter.ac.uk

Taha Mansouri
Senior Lecturer in AI
University of Salford, UK
Email: T.Mansouri@salford.ac.uk

## Abstract

Artificial intelligence (AI) systems increasingly enact market-facing promises through chatbots, recommendation systems, automated decisions, and generative interfaces. Their failures, misuse, and misrepresentation raise a question that conventional brand-crisis models do not fully specify: how do stakeholders assign responsibility when technical causation, customer-facing control, and governance duties are distributed across an AI system, developer, deployer, vendor, and user? This conceptual paper develops a sociotechnical process theory from a structured, federated scoping synthesis of verified academic and primary sources. It distinguishes an AI/algorithmic incident from an AI-related organisational crisis and, in turn, from an AI-related organisational scandal. The framework proposes that incident configuration shapes actor-specific attribution; attribution informs capability, integrity, fairness, and relationship appraisals; and public moralisation may, but need not, escalate an incident into scandal. The theory offers a reconciliation (requiring direct comparative tests) of findings that algorithm involvement can buffer negative brand reactions in some settings while robot and chatbot failures can redirect responsibility to an associated firm in others. It introduces *accountable transparency* as a proposed response configuration that combines timely notice, an intelligible account, role-responsibility acknowledgement, remedy, evidence of correction, and recourse. The evidence supports conditional, proximal inferences about blame, trust, satisfaction, firm evaluation, and communication credibility more strongly than claims about durable reputation, brand equity, or market performance. Accordingly, the paper advances eight explicitly status-coded propositions. The contribution is not a claim that AI invariably intensifies scandal; it is a specification of when AI changes brand-crisis research from a focal-firm problem to a responsibility-network problem.

# 1 Introduction

A chatbot can promise, apologise, recommend, refuse, and mislead in a brand's voice. A recommendation engine can rank products, allocate opportunities, or shape information access without ever appearing human. A developer can supply a model, an integrator can configure it, a professional can rely on it, and a consumer-facing firm can place it at the point of exchange. When an AI-enabled interaction causes harm or violates expectations, each of these actors may become relevant to
stakeholder judgment. The resulting episode cannot be understood solely as a machine error, nor solely as a conventional single-firm crisis. This premise is consistent with calls to integrate psychological, social-political, and technological-structural perspectives in crisis research (Bundy et al., 2017; Pearson & Clair, 1998).

This paper develops a theory of that problem. It asks: *How do AI-related incidents become organisational crises or public scandals, and how do distributed responsibility and response shape their market-facing consequences?* The question matters for marketing because AI systems increasingly become perceptual representatives of brands. They do not merely support back-office operations; they enact promises of accuracy, care, fairness, safety, convenience, and authenticity. Yet the evidence does not support a simple conclusion that AI uniformly amplifies reputational harm. In eight experiments, consumers responded less negatively to algorithm-caused than human-caused brand-harm errors under conditions studied by Srinivasan and Sarial-Abi (2021). In contrast, robot and chatbot service-failure studies report that respondents may assign less responsibility to the automated service provider but more responsibility to the associated firm (Leo & Huh, 2020; Pavone et al., 2023; Ryoo et al., 2024). These results use different technologies, settings, comparison conditions, and judgment targets. They should not be treated as commensurate estimates. They nevertheless motivate a theoretically plausible reconciliation: technical causation by an AI and organisation-directed role responsibility need not be mutually exclusive.

The paper makes four contributions. First, it introduces a nested construct system that separates an AI/algorithmic incident, an AI-related organisational crisis (AI-OC), and an AI-related organisational scandal (AI-OS). This prevents the conflation of technical failure, crisis demand, public moralisation, and durable reputation loss. Second, it proposes a responsibility-network extension to brand-crisis theory. It distinguishes five actor-specific judgments (causal contribution, role responsibility, moral blame, perceived legal duty, and sanction preference) rather than forcing a binary choice between the AI and a human actor.

Third, it specifies a capability–integrity conversion mechanism through which an output error may be reinterpreted as a judgment about organisational character. Fourth, it defines accountable transparency as a future-facing response construct and identifies conditions under which it should be evaluated, rather than claiming that disclosure by itself restores trust.

The contribution is deliberately bounded. The direct literature is strongest for immediate appraisals, including attribution, emotion, satisfaction, trust, firm evaluation, purchase intention, and response-message credibility. Regulatory, tribunal, and company records document incidents, interventions, and formal responses, but normally do not estimate brand equity, sales, employee trust, or durable reputation. The paper therefore treats longer-run brand, legitimacy, and spillover outcomes as proposed outcomes requiring field and longitudinal tests.

# 2 Construct System and Evidence Boundary

## 2.1 A nested account of incident, crisis, and scandal

The central distinction is temporal and interpretive. An *AI/algorithmic incident* is a materially AI-related event, for example, an erroneous, unsafe, biased, offensive, deceptive, or improperly governed output, deployment, or representation. It can be a private service failure and need not involve public attention. An AI-related organisational crisis (AI-OC) arises when an incident creates a material threat to a stakeholder relationship, organisational reputation, or organisational legitimacy and generates a demand for explanation, remedy, correction, or governance response. An

Table 1: Nested construct system for AI-related brand crises

| **Construct** | **Minimum observable criteria** | **Inference boundary** |
|---|---|---|
| AI/algorithmic incident | AI is materially implicated in an output, deployment, use, data practice, or representation that creates actual or credible stakeholder harm. | Does not require public visibility, moralisation, organisational blame, or a reputation effect. |
| AI-related organisational crisis (AI-OC) | An incident threatens a stakeholder relationship, reputation, or legitimacy and creates a demand for organisational account-giving, remedy, | May remain contained; public scandalisation is not required. |
| AI-related organisational scandal (AI-OS) | An AI-OC diffuses beyond directly affected stakeholders and becomes moralised as a violation, producing cross-audience condemnation, institutional demands, or sustained | Does not itself establish durable reputation loss, brand-equity decline, or legal liability. |
| AI-related brand scandal | An AI-OS in which a market-facing brand promise, identity, consumer relationship, or exchange is central. | A subtype of AI-OS; public-sector cases are comparison cases unless a market-facing brand relationship is focal. |
| AI washing | A false, inflated, vague, or misleading representation of AI use, capability, safety, or performance. | May enter directly as an integrity-relevant incident; a malfunctioning AI is not required. |

*Note.* The construct system is proposed for theory development. It distinguishes the object of explanation (incident, crisis, or scandal) from possible outcomes (e.g., trust, firm evaluation, legitimacy, or durable brand equity).

AI-related organisational scandal (AI-OS) is the subset of AI-OCs that crosses a public escalation threshold: the episode diffuses beyond directly affected parties, is morally framed as a violation, and stimulates cross-audience condemnation, institutional demands, or sustained public controversy. Public salience and moralisation are therefore escalation conditions, not conditions that define every AI-related crisis.

This nesting retains two important distinctions. First, an organisation may respond to an AI-OC before it becomes a scandal, and early response can affect whether amplification occurs. Second, public attention is neither a proxy for harm nor evidence of durable brand damage. A viral interaction may be briefly salient without producing lasting consequences; a quieter high-stakes incident may threaten legitimacy even without broad consumer visibility. Table 1 defines the categories and their analytical use.

The technology label is not itself explanatory. "AI" in this paper covers algorithms, machine learning systems, chatbots, robots, generative systems, and other automated decision or interaction systems when their involvement is materially relevant. Probabilism, adaptivity, generativity, embodiment, and autonomy are not defining conditions. They are potentially important moderators because they can influence perceived agency, predictability, controllability, and appropriate reliance (Hoff & Bashir, 2015; J. D. Lee & See, 2004). This avoids implying that a deterministic scoring system, a generative system, and an anthropomorphic service robot produce identical stakeholder responses.

### 2.2 Why an AI-OC is sociotechnical

AI-related incidents are rarely reducible to a model output. They may concern data provenance, a vendor contract, task delegation, interface framing, human review, monitoring, escalation, an unsupported performance claim, or the ability of harmed people to obtain correction. Scholarship on sociotechnical fairness and algorithmic accountability cautions against treating abstract properties of a system as separable from the social and institutional context in which it is deployed (Martin, 2019; Selbst et al., 2019). Algorithmic opacity is likewise not only a technical matter; it can result from the organisation's inability or unwillingness to explain its decisions and responsibilities (Ananny & Crawford, 2018; Burrell, 2016).

This orientation does not deny the relevance of system performance. It changes the unit of analysis from "an error produced by a machine" to "an organisationally attributable episode in a distributed human–AI arrangement." The marketing-specific question is then how that arrangement changes a brand's relationship with its stakeholders. A firm that represents an AI system as competent, caring, fair, safe, or authentic makes an implicit or explicit market-facing promise. The system's conduct may confirm or violate that promise, while governance arrangements help stakeholders infer whether the firm is capable and willing to manage its own technology.

### 2.3 Evidence synthesis and inferential discipline

The framework is grounded in a structured, federated scoping synthesis completed on July - August 2026. Discovery used accessible publisher and DOI records, scholarly search systems, and backward/forward citation chaining. The working search logic combined technology terms (e.g., AI, algorithm, chatbot, robot, generative AI), organisational terms (e.g., brand, firm, company, provider, deployer), incident terms (e.g., failure, bias, deception, privacy, unsafe output, AI washing), and outcome terms (e.g., blame, trust, reputation, legitimacy, crisis response). The initial synthesis map classified evidence as direct AI/organisational evidence, closely related AI evidence, transferred foundational theory, and primary-case evidence.

The review should not be represented as a fully systematic database review. It did not use prospectively registered protocol, independent duplicate screening, or inter-coder reliability assessment. The paper consequently does not claim exhaustive coverage, a PRISMA flow, pooled effects, or prevalence. Appendix A provides the executed conceptual protocol, coding logic, and a source-role register for every reference used in the paper; it does not substitute for a future dual-screened systematic review. Direct evidence is marked [E]; theory transferred from adjacent research is marked [T]; relationships introduced here for empirical assessment are marked [P].

## 3 A Responsibility-Network Process Theory

Figure 1 presents the proposed process theory. An AI-related incident may trigger immediate organisational response and proximal outcomes without becoming publicly scandalised. Scandal escalation is shown as an optional pathway, not an inevitable stage. Longer-run outcomes appear separately because available evidence is thin on their magnitude and persistence. The figure also makes feedback explicit: learning, revised expectations, and regulatory changes can alter subsequent incident configurations.

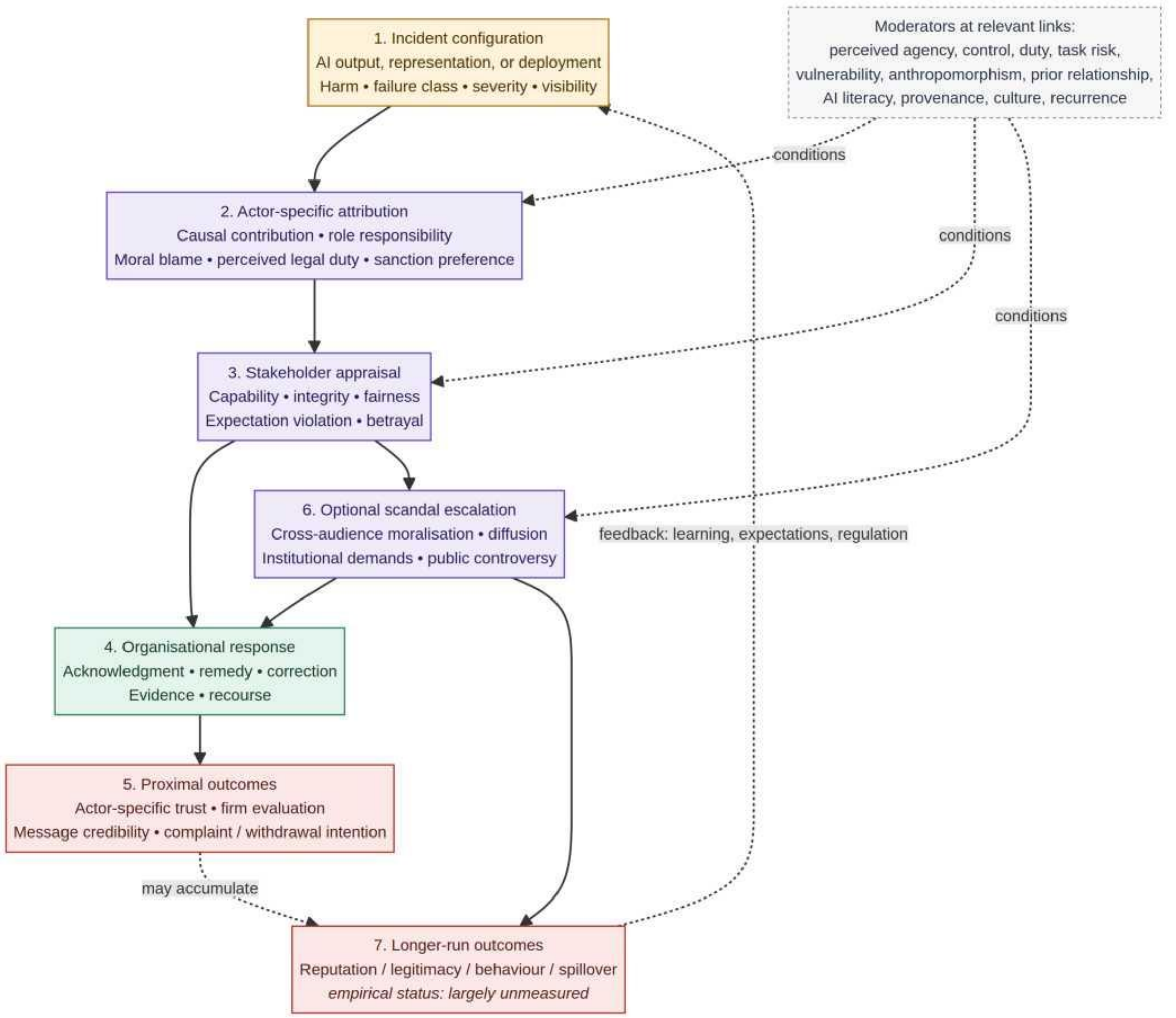

Figure 1: Proposed process model of AI-related organisational crisis and scandal. Solid arrows represent the proposed process; dashed arrows identify conditional or accumulating relationships. "Optional scandal escalation" is not a necessary stage in an AI-related organisational crisis. Longer- run outcomes are theoretically relevant but largely unmeasured in direct AI–brand research.

### 3.1 Stage 1: Incident configuration

The starting point is not a uniform "AI failure" treatment, but an incident configuration. Relevant dimensions include the form of harm; task and industry risk; severity and scale; affected stakeholders and their vulnerability; prior promises; whether the event concerns technical performance, deployment, data governance, or representation; visibility; recurrence; and whether exposure was self-initiated or external. Control-related conditions are also central: who selected the system, configured it, set the task, reviewed output, designed the interface, monitored performance, and can provide redress.

A particularly important conceptual distinction is between predictability of a *specific output* and foreseeability of a *failure class*. A firm might not predict a particular unsupported sentence generated by a language model yet know that unsupported outputs are a foreseeable failure class in a high-stakes setting. This does not imply strict legal liability. It does support a stakeholder attribution question: did the deploying organisation possess a duty and opportunity to guard against a known class of risk?

**Proposition 1 - failure-class foreseeability** [T] [P]. *For a stakeholder evaluating a deploying organisation after an AI-related incident, perceived foreseeability of the relevant failure class will increase perceived organisational role responsibility through perceived preventability, above and beyond the perceived predictability of the particular output. This indirect relationship should be stronger where the task has high duty-of-care implications and the organisation is perceived to have meaningful safeguards or redress control.*

The proposition is a transfer-and-extension claim, not a reported causal result. It builds on appropriate-reliance research (Hoff & Bashir, 2015; J. D. Lee & See, 2004) and accountability reasoning, but needs direct tests that separate known failure classes, output unpredictability, control rights, and stakeholder role.

### 3.2 Stage 2: Actor-specific attribution

Attribution should not be measured as a zero-sum contest between a human and an AI. Five judgments are analytically distinct. *Causal contribution* is the extent to which an observer believes that an actor or system brought about the outcome. *Role responsibility* concerns the obligation to prevent, explain, or remedy harm given an actor's position and control. *Moral blame* is a normative judgment of culpability or wrongdoing. *Perceived legal duty* is a lay inference about legal responsibility and should not be equated with law. *Sanction preference* concerns desired punishment, compensation, regulation, or withdrawal. Each judgment can be separately assessed for the developer, deployer, vendor/integrator, employee or professional user, end user, and perceived AI agent.

The empirical literature reveals why this separation matters. Algorithm-caused errors led to less negative brand response in the experimental settings studied by Srinivasan and Sarial-Abi (2021), with perceived algorithmic agency and responsibility implicated in the process. Three robot-service studies found lower responsibility attributed to a robot, but higher responsibility attributed to the service firm (Leo & Huh, 2020); four further studies report a comparable firm-directed blame shift (Ryoo et al., 2024). In chatbot service-failure experiments, the absence of perceived intention and control in a chatbot was linked to greater company responsibility (Pavone et al., 2023). These are not direct tests of a multi-party developer–deployer–vendor network; they are a basis for theorising that an AI's causal contribution does not exhaust an organisation's role responsibility.

**Proposition 2 - deployer control and role responsibility** [E] [P]. *Following an AI-related incident, perceived control of the deploying organisation over task selection, system configuration, interface presentation, monitoring, and available redress will positively predict perceived role responsibility of that organisation. This relationship will be mediated by perceived preventability and moderated by perceived duty of care; causal contribution attributed to the AI may coexist with, rather than necessarily reduce, deployer role responsibility.*

A legal record illustrates the need to keep domains distinct. In *Moffatt v. Air Canada*, a British Columbia Civil Resolution Tribunal decision found that Air Canada was responsible for misleading information supplied by a website chatbot and ordered a case-specific payment. The decision is a helpful illustration of a customer-facing organisation's legal responsibility in that particular dispute, not evidence of population-level blame or a general chatbot-liability rule (Moffatt v. Air Canada, 2024).

### 3.3 Stage 3: Capability, integrity, and relationship appraisal

Stakeholders may interpret an AI-related incident as evidence about capability (whether an organisation can competently manage the service) or character (whether it is diligent, fair, honest, and responsive). These are related, but not interchangeable, evaluations. The competence–integrity distinction is central in trust-repair research (Gillespie & Dietz, 2009; Kim et al., 2004), while corporate-reputation research separates capability and character reputation (Park & Rogan, 2019). In the AI setting, a model error might initially damage expected capability. Evidence that a firm knowingly delegated an unsuitable task, ignored a known limitation, concealed relevant information, failed to provide recourse, or misrepresented its AI capabilities may create a separate integrity-relevant appraisal.

This proposed conversion should not be read as an assertion that AI always entangles capability and character more than other technologies. The more precise claim is that it identifies conditions under which stakeholders may make the conversion. It also explains why AI washing can enter the model without a technical-output failure. A claim–capability mismatch is itself an integrity relevant incident. Recent marketing scholarship describes a proposed cycle in which irresponsible AI representations and backlash can reinforce mistrust (Ozturkcan & Bozdağ, 2025); the causal trajectory and its market consequences remain open for direct testing.

**Proposition 3 - capability-integrity conversion** [T] [P]. *When a technical AI incident is accompanied by perceived negligent delegation, unaddressed knowledge of a failure class, concealment, lack of recourse, or misleading AI claims, stakeholders will be more likely to infer an integrity violation, rather than a capability shortfall alone. Perceived integrity violation will mediate the association between those governance cues and moral blame of the focal organisation.*

Prior brand associations and relationship contracts are expected to condition this process. Product-harm studies show that prior expectations and response can jointly influence brand outcomes (Dawar & Pillutla, 2000), and prior corporate-responsibility associations can shape attribution and brand evaluation in a product-harm crisis (Klein & Dawar, 2004). Brand-transgression research also shows that effects depend on the relationship contract and brand meaning (Aaker et al., 2004). Analogously, a responsible-AI promise could either provide a positive association or heighten perceived inconsistency when contradicted by an incident. The direction should be tested rather than presumed.

### 3.4 Stage 4: Anthropomorphism and system agency

AI systems can be perceived as social actors, particularly when a chatbot has a name, a face, a conversational style, or a human-like voice. These cues are not simply decorative. In customer– chatbot interactions, anthropomorphism lowered satisfaction, overall firm evaluation, and subsequent purchase intentions among customers who entered the interaction angry; the reported explanation was heightened efficacy expectations and consequent expectancy violation (Crolic et al., 2022). By contrast, a survey-based model of 462 respondents linked anthropomorphism, empathy, and interaction quality to sustained chatbot trust after service failure, with AI anxiety qualifying one relationship (Gu et al., 2024). Studies of AI moral violations also show that people may attribute mind and moral status to AI under some conditions (Shank & DeSanti, 2018).

The literature thus motivates competing, context-sensitive pathways: (a) social presence and empathy can support an interactional recovery pathway; (b) heightened expectations and perceived intentionality can intensify negative appraisal. The model does not predict a universal effect of anthropomorphism.

**Proposition 4 - anthropomorphic double pathway** [E] [P]. *In Ai-mediated service incidents, anthropomorphic interface cues will have two competing indirect associations with evaluation of the deploying brand: a social-presence pathway through perceived empathy and a disappointment pathway through heightened efficacy expectations and perceived intentionality. The relative strength of the disappointment pathway will increase with pre-incident anger, failure severity, and moral framing.*

### 3.5 Stage 5: Organisational response and accountable transparency

An organisational response is part of the evidence stakeholders use to judge the incident. Crisis theory directs attention to situation-appropriate response, reputational threat, and stakeholder reactions (Coombs, 2007); organisation-level trust repair emphasises substantive, multilevel repair rather than communication alone (Gillespie & Dietz, 2009). In AI-related incidents, a technical explanation of how a system works may be important but insufficient. A disclosure that an interface used AI can be interpreted as candour, a warning, a cost-saving substitution, or an attempt to shift risk. AI disclosure effects have been reported as context dependent in marketing contexts (Kirkby et al., 2023; Luo et al., 2019). In crisis communication experiments, credibility and responsibility appraisal were central to acceptance of an AI-scripted response (Ray et al., 2025), while AI-authored apologies received lower perceived sincerity in a study where warmth mitigated part of the indirect effect on trust and forgiveness (Lim et al., 2025).

We propose *accountable transparency* as a formative response configuration, not a synonymous label for disclosure or a demonstrated intervention. It contains six potentially complementary components: (1) timely notice of material AI involvement when relevant; (2) an intelligible account of established facts and uncertainty; (3) acknowledgement of the organisation's role in prevention and remedy without premature assertions about legal fault; (4) immediate, accessible remedy; (5) evidence of technical and governance correction; and (6) a route for contesting a decision, obtaining assistance, or seeking redress. The configuration is distinct from an explanation because it locates responsibility and recourse; it is distinct from an apology because it contains verifiable remedial components.

**Proposition 5 - accountable transparency** [T] [P]. *After an Ai-related organisational crisis, a response configuration comprising role-responsibility acknowledgement, material fact/uncertainty disclosure, immediate remedy, evidence of corrective action, and accessible recourse will be associated*
*with stronger actor-specific trust-repair evaluations than source disclosure alone. This relationship will be mediated by perceived procedural fairness and response credibility and will be stronger when the response matches the perceived violation type.*

The proposition is intentionally not a claim that every component is necessary, sufficient, or equally weighted. It specifies an empirical programme: test individual components, complementarities, timing, costs, and effects across capability- versus integrity-relevant incidents.

**Proposition 6 - response-violation fit** [T] [P]. *For a capability-dominant AI incident, evidence of technical correction and risk calibration will more strongly predict trust-repair evaluations when paired with accurate acknowledgment. For an integrity-dominant AI incident, role-responsibility acknowledgement, remedy, and governance reform will more strongly predict trust-repair evaluations than technical correction alone.*

This is an extension of competence–integrity repair logic rather than advice to deny or admit legal culpability. Appropriate response is contingent on evidence, stakeholder harm, jurisdiction, and the organisation's actual role.

### 3.6 Stage 6: Optional scandal escalation and outcome differentiation

A crisis becomes a scandal only when the episode acquires meaning across audiences. Social amplification theory highlights how information transfer and societal response can amplify or attenuate risk signals (Kasperson et al., 1988); organisational-misconduct research addresses publicised norm violations and their consequences (Greve et al., 2010). In this theory, a proposed scandal outcome is observable as diffusion beyond directly affected parties, moralised condemnation, and demands for institutional accountability or sanction. It is not defined merely by a high number of views or a negative headline.

**Proposition 7 - scandal escalation** [T] [P]. *An AI-related organisational crisis will be more likely to escalate into an AI-related organisational scandal when cross-audience moralisation and diffusion occur together with salient organisational attribution and institutional demands for* *account giving or sanction. Victim vulnerability, absence of meaningful recourse, recurrence, and linkage to an established public risk narrative will strengthen this relationship.*

The proposition is intentionally multilevel. It predicts an escalation outcome at the public discourse level, but it assumes that individual and organisational appraisals are its inputs. Its direction and strength require media-network, institutional, and stakeholder research rather than individual vignettes alone.

Longer-run outcomes must remain disaggregated. Immediate message credibility, actor-specific trust, firm evaluation, complaint, and withdrawal intentions are not equivalent to durable brand equity, market behaviour, institutional legitimacy, or ecosystem spillover. Conventional product harm evidence establishes that category spillover can occur in some markets (Cleeren et al., 2013); there is little direct evidence of developer–deployer or AI-category spillover following AI-related incidents.

**Proposition 8 - governance-linkage spillover** [T] [P]. *Following a publicly escalated AI-related organisational scandal, negative evaluation of a focal organisation will spil l over to an associated developer, deployer, or AI category to the extent that observers perceive technological and governance linkage between them. Perceived shared control, co-brand salience, and response coordination will mediate this relationship.*

## 4 Illustrative Cases and the Limits of Case Inference

The purpose of cases here is not to prove the proposed mechanisms, estimate prevalence, or demonstrate durable reputation effects. Instead, primary and authoritative records make the sociotechnical actor chain observable: they show how an incident can involve a customer interface, safety programme, data practice, vendor relationship, representation, or post-incident account. Each source is used only for the fact type it can establish.

The records do not show a uniform "vendor non-shield" effect. They instead make a more careful conclusion possible: vendor relationships do not remove the need to analyse deployment responsibilities, stakeholder attribution, and the availability of remedy. Nor do cases permit a conclusion that an AI incident inevitably produces durable brand damage. In this paper, those questions remain theoretical outcomes that motivate P2 and P8.

# 5 Implications for Marketing Theory

## 5.1 From a focal firm to a responsibility network

Brand-crisis research has long acknowledged stakeholders, external attributions, and category spillover. The present framework does not claim that conventional theory assumes an isolated firm in every setting. Its more limited contribution is to specify a missing level of granularity in AI-enabled exchange: the technical and governance relationship among developers, deployers, vendors, professional users, end users, and perceived AI agents. These relationships matter not because every actor is equally culpable, but because their allocation of technical capability, decision rights, customer access, information, and redress may cue distinct responsibility judgments.

The theory is most marketing-specific when it considers the brand as a relational promise enacted through AI. A brand-facing AI system can produce two linked but separable relationships: stakeholder–organisation and stakeholder–AI agent. In some contexts, an associated developer is a third organisational relationship. This triadic or multi-actor perspective invites research on the conditions under which trust transfers across relationships, when a chatbot is treated as a representative rather than merely a tool, and when a vendor relationship becomes salient to customers, employees, or professional clients.

## 5.2 From reliable products to governable services

Conventional product-harm response often presupposes that a discrete defect can be recalled, corrected, or removed. An AI service can be updated and controlled, but stochastic outputs and changing deployment contexts make a universal guarantee of error elimination implausible. This suggests a future construct of *governability*: the perceived capacity of an organisation to identify risks, constrain use, document decisions, monitor performance, correct harm, and provide recourse. Governability is not measured or validated here. It is offered as a bridge between technical risk management and stakeholder evaluation, suitable for scale development and comparative research.

## 5.3 From disclosure to accountable transparency

Disclosure alone is an underspecified marketing intervention. It can signal honesty, yet it can also lower expectations of care or invite concern about authenticity. The proposed accountable- transparency configuration shifts the focus from whether AI is mentioned to whether a brand provides an intelligible and challengeable account. It connects source disclosure with procedural fairness, corrective action, and the maintenance of a stakeholder relationship. Future studies should test whether the configuration is formative, whether particular components substitute for one another, and how effects differ when AI involvement is material, legally required, expected, or irrelevant.

## 6 Managerial and Policy Implications

The framework does not establish a universal reputation-repair playbook. It does, however, provide precautionary design questions. Before deployment, organisations can map the full actor and control chain; identify foreseeable failure classes and affected groups; determine when human escalation and recourse are needed; substantiate AI-related claims; and document monitoring, versioning, and vendor responsibilities. These activities are best treated as governance resources and sources of evidence, not as guarantees that a crisis will not occur.

At incident detection, managers should distinguish between technical causation, organisational role responsibility, and legal fault. Prompt acknowledgement of stakeholder needs, preservation of evidence, immediate remedy where possible, and accurate communication of known facts need
not prejudge legal liability or final root cause. The applicable law and appropriate disclosure will vary by jurisdiction, task, contractual setting, and the nature of harm. A response that transfers blame to a vendor or AI system without explaining the organisation's own customer-facing role may create additional integrity concerns; whether it does so is an empirical prediction rather than an established universal effect.

For developers and platforms, the paper highlights the market relevance of downstream governance support: limitation information, context-appropriate instructions, monitoring hooks, incident notification pathways, safe fallbacks, and version traceability can enable a deployer to give stakeholders a more credible account. For policymakers, the theory underscores the importance of clear duties and remedies across multiple actors. It is consistent with, but does not test the effectiveness of, risk-based governance vocabulary such as that in the NIST AI Risk Management Framework and the actor- and risk-contingent structure of the European Union's AI Act. Neither framework is evidence that compliance prevents a crisis or repairs reputation (European Parliament and Council of the European Union, 2024; Tabassi, 2023).

## 7 Limitations and Conclusion

This article is a theory-building synthesis with several limitations. Its federated scoping procedure is not exhaustive and lacked the duplicate screening, formal risk-of-bias assessment, database denominators, and reproducible numerical flow required for a full systematic review. Its direct evidence base is heterogeneous and concentrates on experimental, customer-service, chatbot, and robot settings. Its primary-case illustrations are purposive and selected for conceptual variation, not representativeness. Legal decisions, regulatory actions, settlements, and company accounts have different evidentiary status and cannot be combined as a uniform measure of stakeholder judgment. Finally, the framework's most distinctive claims (failure-class foreseeability, multi-actor responsibility, accountable transparency, scandal escalation, governability, and governance-linkage spillover) require direct validation.

These limitations justify restraint, not inaction. The central claim is not that AI failures are more scandal-prone than ordinary software failures, or that they inevitably erode brand equity. It is that AI changes the architecture of explanation. An output can be causally connected to a model while stakeholders simultaneously judge a brand's control, role, candour, and capacity for remedy. An error can remain a contained service failure, develop into an organisational crisis, or become a public scandal when it is moralised and amplified across audiences. By distinguishing those pathways, marketing scholars can move from asking whether "the AI" or "the brand" is blamed to studying how responsibility networks shape market relationships under technological uncertainty.

# A Structured Federated Scoping Protocol and Source-Role Register

### A.1 Search and screening logic

The conceptual query architecture was: (AI OR artificial intelligence OR generative AI OR LLM OR algorithm OR automation OR machine learning OR chatbot OR recommender system OR robot) AND (brand OR corporate OR organisation OR company OR firm OR service) AND (scandal OR crisis OR failure OR harm OR misconduct OR backlash OR reputation OR trust OR legitimacy OR blame OR accountability OR apology).

Additional topic blocks paired algorithmic systems with responsibility, controllability, agency, service failure, recovery, disclosure, AI washing, privacy, bias, safety, or crisis communication. Items were retained when they contributed either direct evidence involving

AI/algorithms and organisational/brand outcomes, a closely related AI mechanism relevant to stakeholder evaluation, a transferable foundation in brand/crisis/trust/legitimacy theory, or authoritative case documentation. Purely technical benchmark papers without organisational or stakeholder relevance were excluded from the article. Case records were retained when AI was materially implicated, at least one organisation was identifiable, and an authoritative source documented the incident, a response, or legal/regulatory posture. Each source was coded for evidence role, outcome target, design/posture, and safe inferential use.

### A.2 Coding distinction used in the paper

Evidence was separated into direct evidence ([E]), transferred foundation ([T]), and new proposition ([P]). Direct evidence supports only the outcomes actually measured. Transferred sources support a candidate mechanism and are qualified as theoretical foundations. New propositions are distinctive relationships introduced by the paper and must be empirically tested. This coding prevents the paper from using a case record or a product-harm result as proof of a long-run AI-specific reputation effect.

## References

Aaker, J., Fournier, S., & Brasel, S. A. (2004). When good brands do bad. *Journal of Consumer Research*, *31* (1), 1–16. https://doi.org/10.1086/383419

Ananny, M., & Crawford, K. (2018). Seeing without knowing: Limitations of the transparency ideal and its application to algorithmic accountability. *New Media & Society*, *20* (3), 973–989. https://doi.org/10.1177/1461444816676645

Bundy, J., Pfarrer, M. D., Short, C. E., & Coombs, W. T. (2017). Crises and crisis management: Integration, interpretation, and research development. *Journal of Management*, *43* (6), 1661– 1692. https://doi.org/10.1177/0149206316680030

Burrell, J. (2016). How the machine "thinks": Understanding opacity in machine learning algorithms. *Big Data & Society*, *3* (1). https://doi.org/10.1177/2053951715622512

California Department of Motor Vehicles. (2023, October). DMV statement on Cruise LLC suspension. https://www.dmv.ca.gov/portal/news-and-media/dmv-statement-on-cruise-llc-suspension/

Cleeren, K., van Heerde, H. J., & Dekimpe, M. G. (2013). Rising from the ashes: How

brands and categories can overcome product-harm crises. *Journal of Marketing*, *77* (2), 58–77. https://doi.org/10.1509/jm.10.0414

Coombs, W. T. (2007). Protecting organization reputations during a crisis: The development and application of situational crisis communication theory. *Corporate Reputation Review*, *10* (3), 163–176. https://doi.org/10.1057/palgrave.crr.1550049

Crolic, C., Thomaz, F., Hadi, R., & Stephen, A. T. (2022). Blame the bot: Anthropomorphism and anger in customer–chatbot interactions. *Journal of Marketing*, *86* (1), 132–148. https: //doi.org/10.1177/00222429211045687

Dawar, N., & Pillutla, M. M. (2000). Impact of product-harm crises on brand equity: The moderating role of consumer expectations. *Journal of Marketing Research*, *37* (2), 215–226. https : //doi.org/10.1509/jmkr.37.2.215.18729

European Parliament and Council of the European Union. (2024). Regulation (EU) 2024/1689 of the European Parliament and of the Council of 13 june 2024 laying down harmonised rules on artificial intelligence (Artificial Intelligence Act). https://eur-lex.europa.eu/eli/reg/2024/ 1689/oj

Federal Trade Commission. (2024, March). Rite Aid Corporation, FTC v. https://www.ftc.gov/legal-library/browse/cases-proceedings/2023190-rite-aid-corporation-ftc-v

Gillespie, N., & Dietz, G. (2009). Trust repair after an organization-level failure. *Academy of Management Review*, *34* (1), 127–145. https://doi.org/10.5465/amr.2009.35713319

Greve, H. R., Palmer, D., & Pozner, J.-E. (2010). Organizations gone wild: The causes, processes, and consequences of organizational misconduct. *Academy of Management Annals*, *4* (1), 53–107. https://doi.org/10.1080/19416521003654186

Gu, C., Zhang, Y., & Zeng, L. (2024). Exploring the mechanism of sustained consumer trust in AI chatbots after service failures: A perspective based on attribution and CASA theories. *Humanities and Social Sciences Communications*, *11*, 1400. https://doi.org/10.1057/s41599- 024-03879-5

Hoff, K. A., & Bashir, M. (2015). Trust in automation: Integrating empirical evidence on factors that influence trust. *Human Factors*, *57* (3), 407–434. https://doi.org/10.1177/0018720814547570

Huang, K., & Wu, F. (2025). The organization–AI–user responsibility triangle: Public understandings of AI and expectations for organizational responses in AI-service-failure crises. *Emerging Media*, *3* (3), 499–525.

https://doi.org/10.1177/27523543251365451

Huang, K., & Wu, F. (2026). Corrigendum to "the organization–AI–user responsibility triangle: Public understandings of AI and expectations for organizational responses in AI-service- failure crises". *Emerging Media*, *4* (2), 396. https://doi.org/10.1177/27523543261453534

Kasperson, R. E., Renn, O., Slovic, P., Brown, H. S., Emel, J., Goble, R., Kasperson, J. X., & Ratick, S. (1988). The social amplification of risk: A conceptual framework. *Risk Analysis*, *8* (2), 177–187. https://doi.org/10.1111/j.1539-6924.1988.tb01168.x

Kim, P. H., Ferrin, D. L., Cooper, C. D., & Dirks, K. T. (2004). Removing the shadow of suspicion: The effects of apology versus denial for repairing competence- versus integrity-based trust violations. *Journal of Applied Psychology*, *89* (1), 104–118. https://doi.org/10.1037/0021- 9010.89.1.104

Kirkby, A., Baumgarth, C., & Henseler, J. (2023). To disclose or not disclose, is no longer the question—effect of AI-disclosed brand voice on brand authenticity and attitude. *Journal of Product & Brand Management*, *32* (7), 1108–1122. https://doi.org/10.1108/JPBM-02-2022- 3864

Klein, J., & Dawar, N. (2004). Corporate social responsibility and consumers' attributions and brand evaluations in a product-harm crisis. *International Journal of Research in Marketing*, *21* (3), 203–217. https://doi.org/10.1016/j.ijresmar.2003.12.003

Lee, J. D., & See, K. A. (2004). Trust in automation: Designing for appropriate reliance. *Human Factors*, *46* (1), 50–80. https://doi.org/10.1518/hfes.46.1.50_30392

Lee, P. (2016, March). Learning from Tay's introduction. https://blogs.microsoft.com/blog/2016/ 03/25/learning-tays-introduction/

Leo, X., & Huh, Y. E. (2020). Who gets the blame for service failures? attribution of responsibility toward robot versus human service providers and service firms. *Computers in Human Behavior*, *113*, 106520. https://doi.org/10.1016/j.chb.2020.106520

Lim, J. S., Hong, N., & Schneider, E. J. (2025). How warm-versus competent-toned AI apologies affect trust and forgiveness through emotions and perceived sincerity. *Computers in Human Behavior*, *172*, 108761. https://doi.org/10.1016/j.chb.2025.108761

Luo, X., Tong, S., Fang, Z., & Qu, Z. (2019). Frontiers: Machines vs. humans: The impact

of artificial intelligence chatbot disclosure on customer purchases. *Marketing Science*, *38* (6), 937–947. https://doi.org/10.1287/mksc.2019.1192

Martin, K. (2019). Ethical implications and accountability of algorithms. *Journal of Business Ethics*, *160*, 835–850. https://doi.org/10.1007/s10551-018-3921-3

Moffatt v. Air Canada. (2024). 2024 BCCRT 149. https://canlii.ca/t/k2spq

Ozturkcan, S., & Bozdağ, A. A. (2025). Responsible AI in marketing: AI booing and AI washing cycle of AI mistrust. *International Journal of Market Research*, *67* (6), 696–722. https : //doi.org/10.1177/14707853251379285

Park, B., & Rogan, M. (2019). Capability reputation, character reputation, and exchange partners' reactions to adverse events. *Academy of Management Journal*, *62* (2), 553–578. https : //doi.org/10.5465/amj.2016.0445

Pavone, G., Meyer-Waarden, L., & Munzel, A. (2023). Rage against the machine: Experimental insights into customers' negative emotional responses, attributions of responsibility, and coping strategies in artificial intelligence–based service failures. *Journal of Interactive Marketing*, *58* (1), 52–71. https://doi.org/10.1177/10949968221134492

Pearson, C. M., & Clair, J. A. (1998). Reframing crisis management. *Academy of Management Review*, *23* (1), 59–76. https://doi.org/10.5465/amr.1998.192960

Raghavan, P. (2024, February). Gemini image generation got it wrong. we'll do better. https : //blog.google/products-and-platforms/products/gemini/gemini-image-generation-issue/

Ray, E. C., Merle, P. F., & Lane, K. (2025). Generating credibility in crisis: Will an AI-scripted response be accepted? *International Journal of Strategic Communication*, *19* (2), 158–175. https://doi.org/10.1080/1553118X.2024.2435494

Ryoo, Y., Jeon, Y. A., & Kim, W. J. (2024). The blame shift: Robot service failures hold service firms more accountable. *Journal of Business Research*, *171*, 114360. https://doi.org/10.1016/ j.jbusres.2023.114360

Selbst, A. D., boyd danah, d., Friedler, S. A., Venkatasubramanian, S., & Vertesi, J. (2019). Fairness and abstraction in sociotechnical systems. *Proceedings of the Conference on Fairness, Accountability, and Transparency*, 59–68. https://doi.org/10.1145/3287560.3287598

Shank, D. B., & DeSanti, A. (2018). Attributions of morality and mind to artificial intelligence after real-world moral violations. *Computers in Human Behavior*, *86*, 401–

411. https : //doi.org/10.1016/j.chb.2018.05.014

Srinivasan, R., & Sarial-Abi, G. (2021). When algorithms fail: Consumers' responses to brand harm crises caused by algorithm errors. *Journal of Marketing*, *85* (5), 74–91. https://doi.org/10. 1177/0022242921997082

Tabassi, E. (2023). *Artificial intelligence risk management framework (AI RMF 1.0)* (tech. rep. No. NIST AI 100-1). National Institute of Standards and Technology. https://doi.org/10. 6028/NIST.AI.100-1

U.S. Department of Justice, U.S. Attorney's Office, Northern District of California. (2024, November). Cruise admits to submitting a false report to influence a federal investigation and agrees to pay $500,000. https://www.justice.gov/usao-ndca/pr/cruise-admits-submitting-false-report- influence- federal- investigation-and- agrees- pay

U.S. Securities and Exchange Commission. (2024, March). SEC charges two investment advisers with making false and misleading statements about their use of artificial intelligence. https: //www.sec.gov/newsroom/press-releases/2024-36